\documentclass[nohyperref]{article}

\usepackage{microtype}
\usepackage{graphicx}
\usepackage{subfigure}
\usepackage{caption}
\usepackage{booktabs} 
\usepackage{makecell}
\usepackage{multicol}
\usepackage{multirow}
\usepackage{float}
\usepackage{enumitem}
\usepackage{wrapfig}
\usepackage{soul}

\usepackage{hyperref}

\newcommand{\ours}{Ours} 
\newcommand{\din}{\mathcal{D}_{\text{in}}}
\newcommand{\dintest}{\mathcal{D}_{\text{in}}^{\text{test}}}
\newcommand{\dtail}{\mathcal{D}_{\text{tail}}}
\newcommand{\dout}{\mathcal{D}_{\text{out}}}
\newcommand{\douttest}{\mathcal{D}_{\text{out}}^{\text{test}}}
\newcommand{\Lc}{\mathcal{L}_{\text{c}}}
\newcommand{\Lin}{\mathcal{L}_{\text{in}}}
\newcommand{\Lout}{\mathcal{L}_{\text{out}}}

\usepackage[accepted]{icml2022}

\usepackage{amsmath}
\usepackage{amssymb}
\usepackage{mathtools}
\usepackage{amsthm}
\usepackage{pifont}
\newcommand{\cmark}{\ding{51}}%
\newcommand{\xmark}{\ding{55}}%

\usepackage[capitalize,noabbrev]{cleveref}

\theoremstyle{plain}

\theoremstyle{definition}

\theoremstyle{remark}

\usepackage[textsize=tiny]{todonotes}

\icmltitlerunning{
 Partial and Asymmetric Contrastive Learning for  Out-of-Distribution Detection in Long-tailed Recognition
}

\begin{document}

\twocolumn[
\icmltitle{
Partial and Asymmetric Contrastive Learning for \\ Out-of-Distribution Detection in Long-Tailed Recognition
}



\icmlsetsymbol{equal}{*}

\begin{icmlauthorlist}
\icmlauthor{Haotao Wang}{utaustin}
\icmlauthor{Aston Zhang}{amazon}
\icmlauthor{Yi Zhu}{amazon}
\icmlauthor{Shuai Zheng}{amazon}
\icmlauthor{Mu Li}{amazon}
\icmlauthor{Alex Smola}{amazon}
\icmlauthor{Zhangyang Wang}{utaustin}
\end{icmlauthorlist}

\icmlaffiliation{utaustin}{University of Texas at Austin, Austin, USA}
\icmlaffiliation{amazon}{Amazon Web Services, Santa Clara, USA}

\icmlcorrespondingauthor{Haotao Wang}{\href{mailto:htwang@utexas.edu}{htwang@utexas.edu}}
\icmlcorrespondingauthor{Aston Zhang}{\href{mailto:astonz@amazon.com}{astonz@amazon.com}}
\icmlcorrespondingauthor{Zhangyang Wang}{\href{mailto:atlaswang@utexas.edu}{atlaswang@utexas.edu}}

\icmlkeywords{Out-of-distribution detection, long-tailed recognition, contrastive learning}

\vskip 0.3in
]



\printAffiliationsAndNotice{Work done during the first author's internship at Amazon Web Services.} 

\begin{abstract}
Existing out-of-distribution (OOD) detection methods are typically benchmarked on training sets with balanced class distributions. However, in real-world applications, it is common for the training sets to have long-tailed distributions.
In this work, we first demonstrate that existing OOD detection methods commonly suffer from significant performance degradation when the training set is long-tail distributed.
Through analysis, we posit that this is because the models struggle to distinguish the minority tail-class in-distribution samples from the true OOD samples, 
making the tail classes more prone to be falsely detected as OOD.
To solve this problem, we propose \textit{Partial and Asymmetric Supervised Contrastive Learning} (PASCL), which explicitly encourages the model to distinguish between tail-class in-distribution samples and OOD samples. 
To further boost in-distribution classification accuracy, we propose Auxiliary Branch Finetuning, which uses two separate branches of BN and classification layers for anomaly detection and in-distribution classification, respectively.
The intuition is that in-distribution and OOD anomaly data have different underlying distributions.
Our method outperforms previous state-of-the-art method by $1.29\%$, $1.45\%$, $0.69\%$ anomaly detection false positive rate (FPR) and $3.24\%$, $4.06\%$, $7.89\%$ in-distribution classification accuracy on CIFAR10-LT, CIFAR100-LT, and ImageNet-LT, respectively. Code and pre-trained models are available at \url{https://github.com/amazon-research/long-tailed-ood-detection}.
\end{abstract}

\section{Introduction}
\label{sec:intro}

Deep neural networks (DNNs) are widely known to be overconfident on what they do not know. Given a test sample that does not belong to any training class,
DNNs tend to recklessly predict this out-of-distribution (OOD) test sample as one of the training classes with high confidence, which is doomed to be wrong \cite{hendrycks2016baseline,hein2019relu}.
Numerous OOD detection methods have been proposed to solve this problem with promising results \cite{hendrycks2018deep,mohseni2020self,liu2020energy}.
The goal of OOD detection is two-fold: (i) to distinguish OOD samples from  in-distribution  samples (i.e., \textit{anomaly detection}), and (ii) to still achieve high accuracy on the \textit{in-distribution sample classification}.

Existing OOD detection methods are typically evaluated on balanced training sets. However, the training sets in real-world applications often follow a long-tailed distribution \cite{van2017devil,liu2019large,cui2019class,menon2020long}. 
In view of this, we benchmark existing OOD detection methods (including both widely used classic methods and recently published ones) on the long-tailed training sets. We observe significant performance drop in terms of both anomaly detection and in-distribution classification, when compared with the results obtained on the balanced training sets (Table \ref{tab:benchmark-ood}). 
This shows the challenge of OOD detection on long-tailed recognition (LTR) task.
Moreover, we show that it cannot be easily fixed by naive solutions such as combining state-of-the-art OOD detection and long-tailed recognition methods (Table \ref{tab:benchmark-ood-lt}). 
Through feature visualization (Fig. \ref{fig:scatter}), we further unveil the underlying cause for such performance drop: The model struggles to distinguish tail-class in-distribution samples from OOD samples. 
In contrast, the head-class in-distribution samples can be easily distinguished from OOD samples.

Motivated by such observations, we propose \textit{Partial and Asymmetric Supervised Contrastive Learning} (PASCL) as the solution for OOD detection in long-tailed recognition. 
The key intuition is to push tail-class in-distribution samples away from OOD samples in the feature space through supervised contrastive learning (SCL).
The \textbf{core idea} of our technical design is that head-class in-distribution, tail-class in-distribution, and OOD samples play different roles in long-tailed OOD detection, and thus they should be treated differently in contrastive learning. Compared to traditional supervised contrastive learning \cite{khosla2020supervised}, PASCL is featured by two important insights tailored for OOD detection on long-tailed training sets: \textit{partiality} and \textit{asymmetry}.

\vspace{-1em}
\paragraph{$\rhd$ Partiality:} Through feature visualization (Fig. \ref{fig:scatter}), we observe tail-class in-distribution samples to be heavily overlapped with OOD samples in the feature space, while head-class in-distribution samples are relatively well separated from OOD samples. 
For this reason, we only partially apply contrastive loss on tail-class in-distribution and OOD samples, rather than on all training samples. In other words, we only push OOD data away from tail-class in-distribution data. 
Otherwise, if we simultaneously pushed OOD samples away from all in-distribution samples, the learning process would bias towards discriminating the head-class in-distribution and OOD data, so that the difference between tail-class in-distribution and OOD data could not be effectively learned. This is because head-class in-distribution and OOD data make up the great majority of the training set.
Experimental results show partiality plays an important role in OOD detection under long-tailed recognition.

\vspace{-1em}
\paragraph{$\rhd$ Asymmetry:} For traditional supervised contrastive learning, samples from the same classes are considered as positive pairs. However, in OOD detection, the training OOD images are not necessarily from the same class, nor should they be pulled together in the feature space. Indeed, the OOD training set usually has vast diversity (in both low-level visual features and high-level semantics) in order to be representative enough for the open visual world. 
Therefore, while pushing tail-class in-distribution samples away from OOD samples, we pull tail-class in-distribution samples within the same classes together, but do \emph{not} pull OOD training samples together. 
Experimental results show that such asymmetric push-and-pull operation on in-distribution and OOD samples is an important knob for good performance. 

To further boost the in-distribution classification accuracy in OOD detection, we propose Auxiliary Branch Finetuning (ABF).
Current state-of-the-art OOD detection methods train the models using both in-distribution and OOD data \cite{hendrycks2018deep,liu2020energy,mohseni2020self}, where the batch normalization (BN) statistics are estimated using both in-distribution and OOD data.
Although such BN statistics of \textit{mixture} distribution between in-distribution and OOD data are beneficial for anomaly detection, they are not the optimal choice for in-distribution classification. 
This is because the in-distribution and OOD data have different underlying distributions, so that the statistics of the mixture distribution estimated in BN do not match those of the in-distribution test data. 
Previous work has shown that even slight mismatch in BN statistics can lead to significant performance decay under covariate shifts \cite{chang2019domain,benz2021revisiting,xie2019intriguing}.
Moreover, the classification layer (i.e., the last fully connected layer) has been empirically shown to play an important role in long-tailed recognition \cite{kang2019decoupling}.
Therefore, 
our auxiliary branch finetuning scheme uses
two separate branches of BN and classification layers for anomaly detection and in-distribution classification, 
while all the other layers are shared across the two tasks. 

We evaluate our method on the recently published semantically coherent OOD detection benchmarks \cite{yang2021semantically}.
PASCL outperforms previous state-of-the-art method by $1.29\%$, $1.45\%$, $0.69\%$ anomaly detection false positive rate (FPR) 
and $3.24\%$, $4.06\%$, $7.89\%$ in-distribution classification accuracy on CIFAR10-LT, CIFAR100-LT \cite{cao2019learning}, and ImageNet-LT \cite{liu2019large}, respectively.

\section{Related Work}
\label{sec:related-works}

\paragraph{OOD detection}
\citet{hendrycks2016baseline} formally studied the OOD detection problem in deep learning. The authors proposed to use the maximum softmax probability (MSP) as a naive baseline for OOD detection, and observed DNNs are commonly overconfident on OOD test samples.  
Other OOD detection measures such as Gram matrix \cite{sastry2020detecting}, Mahalanobis distance \cite{lee2018simple}, and free energy \cite{liu2020energy} were also studied. 

\citet{hendrycks2018deep} proposed the classical Outlier Exposure (OE) approach, which utilized unlabeled auxiliary training set as OOD training data. OE assigns uniform distribution as pseudo labels for OOD training data, and then minimizes the cross-entropy loss on both in-distribution and OOD training data. The use of auxiliary OOD traing sample greatly boosts OOD detection performance compared with previous methods that only use in-distribution training samples. 
OE has since become a cornerstone where many following works are based upon. EnergyOE \cite{liu2020energy} maximizes the free-energy of OOD training samples.  
OECC \cite{PAPADOPOULOS2021138} replaced the cross-entropy loss in OE with total variance loss and further added a confidence-calibration loss term. 
NTOM \cite{chen2021atom} proposed to classify all OOD samples into one abstaining class, and used hard-sample mining to select the most informative OOD training samples in each epoch.

Lately, \citet{yang2021semantically} considered a more challenging scenario where the unlabeled auxiliary training set may contain both in-distribution and OOD data. They proposed unsupervised dual grouping (UDG) to split the unlabeled in-distribution and OOD training samples.
The authors also pointed out that in-distribution data even widely exist in previous OOD benchmarks. 
To solve this problem, the authors constructed a new OOD detection benchmark, named semantically coherent out-of-distribution detection (SC-OOD), which appears more challenging than previous benchmarks \cite{yang2021semantically}.

Several previous works have shown that contrastive learning (CL) and distance metric learning benefits OOD detection on balanced training sets \cite{winkens2020contrastive,zhou2021contrastive,sohn2020learning,yang2020out}.
Our PASCL differs from them since it is tailored for the OOD detection problem on long-tailed dataset, which is a new challenge unaddressed in the prior literature.

\vspace{-1em}
\paragraph{Long-tailed recognition}
The most straight forward solution for long-tailed recognition is to re-balance the training set using undersampling or oversampling \cite{he2009learning}. 
However, such methods lead to unsatisfactory performance in large-scaled deep learning \cite{cui2019class,wang2020long}.
\citet{kang2019decoupling} observed that after the DNN has been trained on imbalanced data, simply retraining the classification layer on a re-balanced dataset can significantly boost long-tailed recognition performance. 
Recently, \citet{menon2020long} proposed a formal statistical framework for long-tailed recognition and a statistically grounded long-tailed recognition method termed logit adjustment that achieves impressive improvements over previous methods. 

\vspace{-1em}
\paragraph{Long-tailed OOD detection}
\citet{roy2022does} studied a long-tail OOD detection problem in medical image analysis, whose problem setting is relevant but intrinsically different with ours. In their work, all the head-classes are in-distribution while all the tail-classes are considered as OOD. Their goal is to distinguish the head-class in-distribution samples from the tail-class OOD samples. In contrast, in our \emph{more general} problem setting, the in-distribution samples come from both head-classes and tail-classes, and the same goes for the OOD samples. 
OLTR \cite{liu2019large} is an long-tailed recognition method taking open set classification into consideration. Its main goal is to improve in-distribution classification accuracy in long-tailed recognition instead of OOD detection. Although it achieves better OOD detection performance than naive baselines such as MSP, it is not comparable with state-of-the-art OOD detection methods such as OE (see Appendix \ref{sec:appx-oltr} for results). In contrast, our paper focuses on improve OOD detection performance on long-tailed training sets. 
More related works are discussed in Appendix \ref{sec:appx-related}.



\section{The Challenge of OOD Detection in Long-Tailed Recognition}
\label{sec:challenge}

In this section, we first examine the performance of existing OOD detection methods on long-tailed training sets.
Specifically, we benchmark six different OOD detection methods, including both widely used classical methods (e.g., MSP \cite{hendrycks2016baseline}, OE \cite{hendrycks2018deep}, EnergyOE \cite{liu2020energy}, SOFL \cite{mohseni2020self} and very recently published new methods (e.g., OECC \cite{PAPADOPOULOS2021138}, NTOM \cite{chen2021atom}). 
We train all methods on CIFAR10 \cite{krizhevsky2009learning} and CIFAR10-LT (i.e., the long-tailed version of CIFAR10) with imbalance ratio\footnote{Imbalance ratio: the number of training samples in the most frequent class divided by that of the least frequent class.} $\rho=100$.

We then evaluate OOD detection performance on the recently published SC-OOD benchmarks.
The results are summarized in Table \ref{tab:benchmark-ood}.\footnote{The evaluation measures will be described in Section \ref{sec:settings}.} 
Compared with the results obtained on the balanced CIFAR10 dataset, CIFAR10-LT leads to significant performance drop in terms of both anomaly detection and in-distribution classification. 

\begin{table}[ht]
\centering
\setlength{\tabcolsep}{3pt}
\vspace{-0.5em}
\caption{The challenge of OOD detection in long-tailed recognition: existing OOD detection methods suffer significant performance drop when trained on long-tailed datasets. For each method, we train two ResNet18 models on CIFAR10 and CIFAR10-LT, respectively. Reported are the average performance across six different OOD test sets in the SC-OOD detection benchmark. 
The performance drop of each method from CIFAR10 training set to CIFAR10-LT training set are shown in parenthesis.
All results are shown in percentage.}
\resizebox{1\linewidth}{!}{
\begin{tabular}{ c|c|cccc }
\toprule
\textbf{Method} & \textbf{Dataset} & \textbf{AUROC ($\uparrow$)} & \textbf{AUPR ($\uparrow$)} & \textbf{FPR95 ($\downarrow$)} & \textbf{ACC ($\uparrow$)} \\
\midrule
\midrule
\multirow{2}{*}{\makecell{NT\\(MSP)}} & CIFAR10 & 85.86 & 84.37 & 52.52 & 93.45 \\
& CIFAR10-LT & 72.28 (-13.58) & 70.27 (-14.10) & 66.07 (+13.55) & 72.34 (-21.11) \\
\midrule
\multirow{2}{*}{OE} & CIFAR10 & 96.68 & 96.29 & 14.59 & 92.81 \\
& CIFAR10-LT & 89.92 (-6.75) & 87.71 (-8.58) & 34.80 (+20.21) & 73.30 (-19.51) \\
\midrule
\multirow{2}{*}{EnergyOE} & CIFAR10 &  96.59 & 96.37 & 14.80 & 93.07 \\
& CIFAR10-LT & 89.31 (-7.27) & 88.92 (-7.45) & 40.88 (+26.08) & 74.68 (-18.39) \\
\midrule
\multirow{2}{*}{SOFL} & CIFAR10 & 96.74 & 96.60 & 14.57 & 89.13 \\
& CIFAR10-LT & 91.13 (-5.61) & 90.49 (-6.10) & 34.98 (+20.41) & 54.42 (-34.71) \\
\midrule
\multirow{2}{*}{OECC} & CIFAR10 & 96.27 & 95.41 & 14.77 & 91.95 \\
& CIFAR10-LT & 87.28 (-8.99) & 86.29 (-9.12) & 45.24 (+30.47) & 60.16 (-31.79) \\
\midrule
\multirow{2}{*}{NTOM} & CIFAR10 & 96.92 & 96.95 & 14.95 & 91.44 \\
& CIFAR10-LT & 92.89 (-4.03) & 92.31 (-4.65) & 29.03 (+14.09) & 66.41 (-25.03) \\
\midrule
\bottomrule
\end{tabular}
}
\label{tab:benchmark-ood}
\end{table}

\begin{table}[ht]
\centering
\setlength{\tabcolsep}{2.8pt}
\vspace{-0.5em}
\caption{Naive combinations of state-of-the-art OOD detection and long-tailed recognition methods cannot address the challenge of OOD detection in long-tailed recognition (LTR).
All methods are trained on CIFAR10/100-LT using ResNet18 and evaluated on the SC-OOD benchmarks. 
The best results are shown in bold and the second-best ones are underlined.
Mean and standard deviation over six random runs are shown for our method.}
\resizebox{1\linewidth}{!}{
\begin{tabular}{ c|ccc|cccc }
\toprule
$\din$ & \textbf{\makecell{OOD\\Detection\\Method}} & & \textbf{\makecell{LTR\\Method}}
& \textbf{AUROC ($\uparrow$)} & \textbf{AUPR ($\uparrow$)} & \textbf{FPR95 ($\downarrow$)} & \textbf{ACC ($\uparrow$)} \\
\midrule
\midrule
\multirow{5}{*}{\makecell{CIFAR\\10-LT}} & \multirow{4}{*}{OE} & \multirow{4}{*}{+} & None & \underline{89.92} & \underline{87.71} & 34.80 & 73.30 \\
& & & Re-weighting & 89.34 & 86.39 & 37.09 & 70.35 \\
& & & $\tau$-norm &  89.58 & 85.88 & \underline{33.80} & 73.33 \\
& & & LA & 89.46 & 86.39 & 34.94 & \underline{73.93} \\
\cline{2-8}
& \multicolumn{3}{c|}{Our method} & \textbf{90.99}$\pm$0.19 & \textbf{89.24}$\pm$0.34 & \textbf{33.36}$\pm$0.79 & \textbf{77.08}$\pm$1.01\\ 
\midrule
\midrule
\multirow{5}{*}{\makecell{CIFAR\\100-LT}} & \multirow{4}{*}{OE} & \multirow{4}{*}{+} & None & 72.62 & \underline{66.73} & 68.69 & 39.33 \\
& & & Re-weighting & 72.07 & 66.05 & 70.62 & 39.42 \\
& & & $\tau$-norm & \underline{72.71} & 66.59 & \underline{68.04} & 40.87 \\
& & & LA & 72.56 & 66.48 & {68.24} & \underline{42.06} \\
\cline{2-8}
& \multicolumn{3}{c|}{Our method} &  \textbf{73.32} $\pm$ 0.32 & \textbf{67.18} $\pm$ 0.10 & \textbf{67.44} $\pm$ 0.58 & \textbf{43.10} $\pm$ 0.47 \\ 
\midrule
\bottomrule
\end{tabular}
}
\label{tab:benchmark-ood-lt}
\end{table}

We further show that such challenge can not be easily solved by naive solutions such as combining state-of-the-art OOD detection methods with state-of-the-art long-tailed recognition methods.
Specifically, we combine OE with three popular long-tailed recognition methods: re-weighting \cite{cui2019class}, $\tau$-norm \cite{kang2019decoupling}, and logits adjustment (LA) \cite{menon2020long}.
The results are shown in Table \ref{tab:benchmark-ood-lt}.
As we can see, combining OE with long-tailed recognition methods does not necessarily bring benefit compared with the original OE baseline: E.g., on CIFAR10-LT, OE has the best AUROC and AUPR compared with all three combination methods. 
In contrast, our method (to be introduced later) achieves considerable improvements.

\begin{figure}[ht]
\centering
\includegraphics[width=0.8\columnwidth]{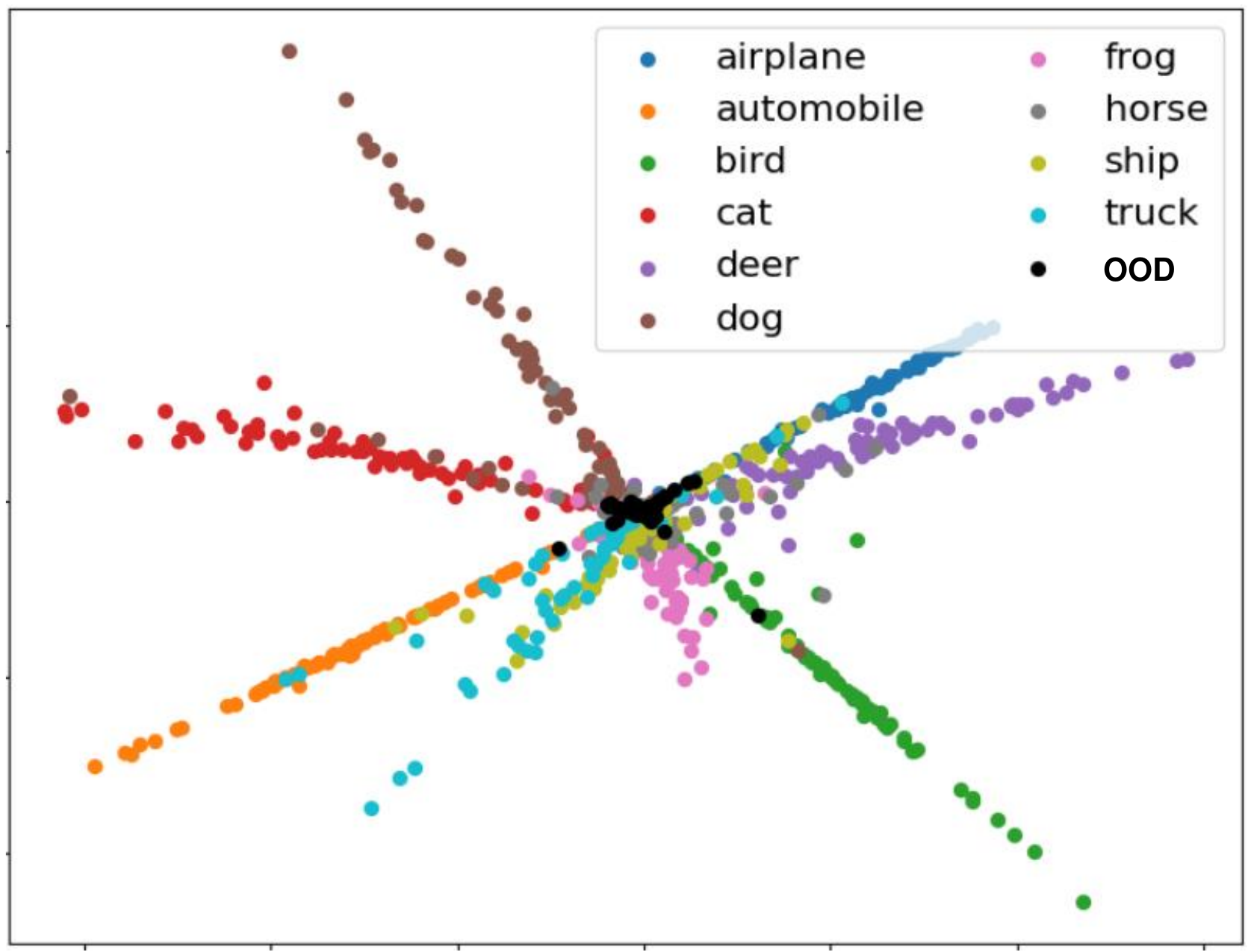}
\vspace{-0.5em}
\caption{Visualization of in-distribution and OOD samples in the feature space.
We visualize the penultimate layer features of ResNet18 trained by OE on CIFAR10-LT, using the same visualization method as in \cite{pang2019rethinking}.
$100$ test samples from each in-distribution class and $100$ OOD samples are visualized.
The model struggles to distinguish tail-class in-distribution classes like ``horse" (grey) and ``ship" (yellow) from OOD samples (black).
Note that in CIFAR10-LT, the number of training samples in each class follows the alphabetic order: the ``airplane'' class has the most training samples and the ``truck'' class has the least.
For comparison, see visualization on the balanced CIFAR10 training set in Appendix \ref{sec:appx-vis}.
}
\vspace{-0.5em}
\label{fig:scatter}
\end{figure}

\vspace{-1em}
\paragraph{Why OE fails on long-tailed recognition}
Feature distributions of models trained by OE on CIFAR10 and CIFAR10-LT are visualized in Fig. \ref{fig:appx-scatter} (in Appendix) and \ref{fig:scatter}, respectively.
As we can see, 
on the balanced CIFAR10 dataset, the OOD samples are well separated from all in-distribution classes (Fig. \ref{fig:appx-scatter}).
In contrast, when trained on CIFAR10-LT, the features of tail-class in-distribution (e.g., horse, ship) samples and OOD samples heavily overlap, while those of head-class in-distribution (e.g., airplane, automobile, bird) samples are well separated from OOD samples (Fig. \ref{fig:scatter}). 
This indicates that on long-tailed datasets, models trained by OE struggles to distinguish tail-class in-distribution samples from OOD samples, which is likely to be a major contributing cause of the performance drop of OE on long-tailed recognition.  
An intuitive explanation for this phenomenon is that the tail-class in-distribution samples have few occurrence in the training set, and are thus more likely to result in low-confidence predictions like the OOD samples.

\section{Method}
\label{sec:solutions}

\subsection{Preliminary}

State-of-the-art OOD detection methods train the model on both in-distribution and OOD training data. 
Let $\din$ and $\dout$ denote an in-distribution training set and an unlabeled OOD training set, respectively. The training objective of the existing OOD detection methods (e.g., OE, EnergyOE, OECC, NTOM, etc.) is formulated as:
\vspace{-0.5em}
\begin{equation}
    \mathcal{L} = 
    \mathbb{E}_{x\sim\din}[\Lin(x)] + 
    \lambda\mathbb{E}_{x\sim\dout}[\Lout(x)],
\label{eq:oe-loss}
\end{equation}
where $\Lin$ is for in-distribution classification and $\Lout$ is to assign low prediction confidence to OOD data.
For example in OE, $\Lin=\text{XEntropy}(x,y(x))$ is the cross-entropy loss on in-distribution data $x$ and the corresponding label $y(x)$, and $\Lout(x)=\text{KL}(f(x)\|u)$ is the KL divergence between the model output (i.e., logits) $f(x)$ and the uniform distribution vector $u$. 

\subsection{Partial and Asymmetric Contrastive Learning}

\begin{figure}[ht]
\centering
\setlength{\tabcolsep}{0.2pt}
\begin{tabular}{c|c}
\includegraphics[width=0.49\columnwidth]{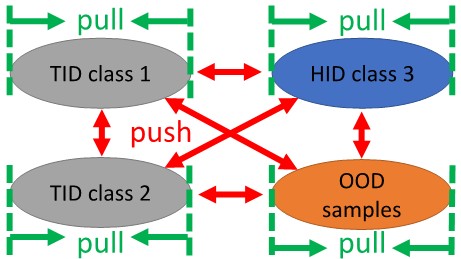} &
\includegraphics[width=0.49\columnwidth]{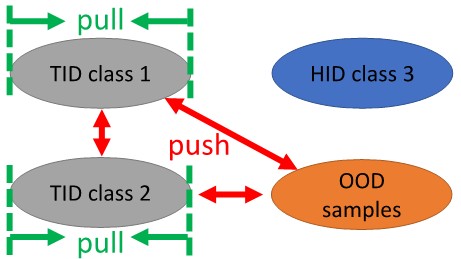} \\
\small{(a) The original SCL} & \small{(b) Our PASCL} \\
\end{tabular}
\caption{Illustration of our partial and asymmetric push-and-pull strategy, in contrast to those in the original supervised contrastive learning (SCL).
The ellipses with different colors represent different types of training samples. Grey: tail-class in-distribution classes; blue: head-class in-distribution classes; orange: OOD training samples.
The green and red arrows show pulling and pushing operations in contrastive learning, respectively.
}
\label{fig:demo}
\end{figure}

Figure \ref{fig:demo} illustrates the core idea of our 
partial and asymmetric supervised contrastive learning 
strategy for OOD detection in long-tailed recognition. 
Traditional supervised contrastive learning treats all classes equally: pulling is applied within each class and pushing between any two different classes (Figure \ref{fig:demo} (a)).
In contrast, PASCL treats head-class in-distribution, tail-class in-distribution, and OOD classes differently:
PASCL applies pulling only within each tail-class in-distribution class, and applies pushing among tail-class in-distribution and OOD samples~(Figure \ref{fig:demo} (b)).

Next we formularize our method to achieve the above partial and asymmetric push-and-pull strategy. 
Specifically, we design the following partial and asymmetric supervised contrastive learning (PASCL) loss: 
\vspace{-0.5em}
\begin{align}
& \Lc = \mathbb{E}_{x\in\mathcal{I}} [\Lc(x)], \text{  where  } \Lc(x) = \nonumber \\
& \sum_{p\in\mathcal{P}(x)} 
    -\frac{1}{|\mathcal{P}(x)|}
    \log\frac
    {\exp(z(x)^{T}z(p)/\tau)}
    {\sum\limits_{a\in\mathcal{A}(x)}\exp(z(x)^{T}z(a)/\tau)},
\label{eq:loss}
\end{align}
where $z(x)$ is the output of a non-linear projection on the model's penultimate layer at sample $x$ (as in the original supervised contrastive learning), $\tau$ is the scaling temperature, 
$\mathcal{P}(x)=\{p|y(p) = y(x), p \in \din\setminus \{x\}\}$ is the set of positive samples with respect to  $x$ (i.e., the samples with identical ground truth labels with $x$), 
$\mathcal{A}(x)$ is the set of all contrastive samples with respect to $x$, 
and $\mathcal{I}$ is the set of training samples where the PASCL loss is applied on.  
At first glance, Eq. (\ref{eq:loss}) \emph{seems} identical with the original supervised contrastive learning loss \cite{khosla2020supervised}, previously used to learn good features for in-distribution classification.
However, the \emph{key difference} lies in the definitions of $\mathcal{I}$ and $\mathcal{A}(x)$, as we detail below.

\vspace{-0.8em}
\paragraph{The original supervised contrastive learning \cite{khosla2020supervised}}
All classes play the same role in the original supervised contrastive learning: Pulling is done within every class and pushing between every pair of different classes. 
Such uniform contrastive loss can be applied either only on in-distribution samples 
(i.e., $\mathcal{I}=\din$, $\mathcal{A}(x)=\din \setminus \{x\}$)
or on both in-distribution and OOD samples
(i.e., $\mathcal{I}=\din \cup \dout$,  $\mathcal{A}(x)=\din \cup \dout \setminus \{x\}$).
In the latter case, all OOD samples are assumed to form the
($C$+1)$^{\mathrm{th}}$
class along with the $C$ in-distribution classes.
Experimental results show the latter empirically achieves slightly better performance than the former when combined with OE (Appendix \ref{sec:appx-original-scl}).

\vspace{-0.8em}
\paragraph{Partial supervised contrastive learning} Here we apply supervised contrastive learning only on tail-class in-distribution and OOD data:
$\mathcal{I}=\dtail \cup \dout$, $\mathcal{A}(x)=\dtail \cup \dout \setminus \{x\}$, where $\dtail$ is the set of all tail-class in-distribution training data.
This design is motivated by our observations in Figure \ref{fig:scatter},
where head-class in-distribution samples are easier to be separated from the OOD samples, since there are enough occurrence of them in the training set. Thus, we do not use an extra supervised contrastive learning loss to explicitly push them away. 
Otherwise, if we simultaneously pushed OOD samples away from all in-distribution training samples, the difference between head-class in-distribution and OOD data would be over-emphasized, hindering the learning of the difference between tail-class in-distribution and OOD samples.

\vspace{-0.8em}
\paragraph{Asymmetric supervised contrastive learning} In this case, OOD training samples are only contained in $\mathcal{A}(x)$ but not in $\mathcal{I}$ :
$\mathcal{I}=\din$, $\mathcal{A}(x)=\din \cup \dout \setminus \{x\}$.
That is, we still push in-distribution samples away from OOD samples and in-distribution samples from different classes, and pull in-distribution samples within the same class together, but do not pull OOD samples together. 
This is because the OOD training samples are not necessarily from the same class. Hence they should not be pulled together in the feature space, as the OOD training set typically has huge diversity in order to be representative for the open visual world.  

\vspace{-0.8em}
\paragraph{Putting them together: the PASCL framework}
Combining partiality and asymmetry together, we have
\vspace{-0.5em}
\begin{equation}
\mathcal{I}=\dtail, \mathcal{A}(x)=\dtail \cup \dout \setminus \{x\}.
\label{eq:set}
\end{equation}

In summary, our final loss function is
\vspace{-0.5em}
\begin{align}
    \mathcal{L} = 
    \mathbb{E}_{x\sim\din}[\Lin(x)] & + 
    \lambda_1\mathbb{E}_{x\sim\dout}[\Lout(x)]  \nonumber \\ 
    & + \lambda_2\mathbb{E}_{x\sim\mathcal{I}}[\Lc(x)],
\label{eq:final}
\end{align}
where $\Lin(x)$ and $\Lout(x)$ are identical with those in OE (see Eq.~\eqref{eq:oe-loss}), and $\Lc(x)$ is as defined in Eq. (\ref{eq:loss}) with $\mathcal{I}$ and $\mathcal{A}(x)$ defined in Eq. (\ref{eq:set}).

\subsection{Auxiliary Branch Finetuning (ABF)}

As demonstrated in Eq. \eqref{eq:oe-loss}, current state-of-the-art OOD detection methods train the models using both in-distribution and OOD data \cite{hendrycks2018deep,liu2020energy,mohseni2020self}, where the BN statistics are estimated using both in-distribution and OOD data.
Such BN statistics of \textit{mixture} distribution between in-distribution and OOD data are beneficial for anomaly detection (i.e., to distinguish in-distribution from OOD samples), but not optimal for in-distribution classification. 
The ideal case is to use mixture BN statistics for anomaly detection and \textit{pure} in-distribution BN statistics for in-distribution classification.
Moreover, the classification layer (i.e., the last fully connected (FC) layer) has been empirically shown to play an important role in long-tailed recognition \cite{kang2019decoupling}.
Thus, it may also be beneficial to use a separate classification layer trained only on $\din$ for better in-distribution classification. 
To this end, we propose Auxiliary Branch Finetuning (ABF).

\begin{figure}[ht]
\centering
\resizebox{1\linewidth}{!}{
\begin{tabular}{c}
     \includegraphics[width=\columnwidth]{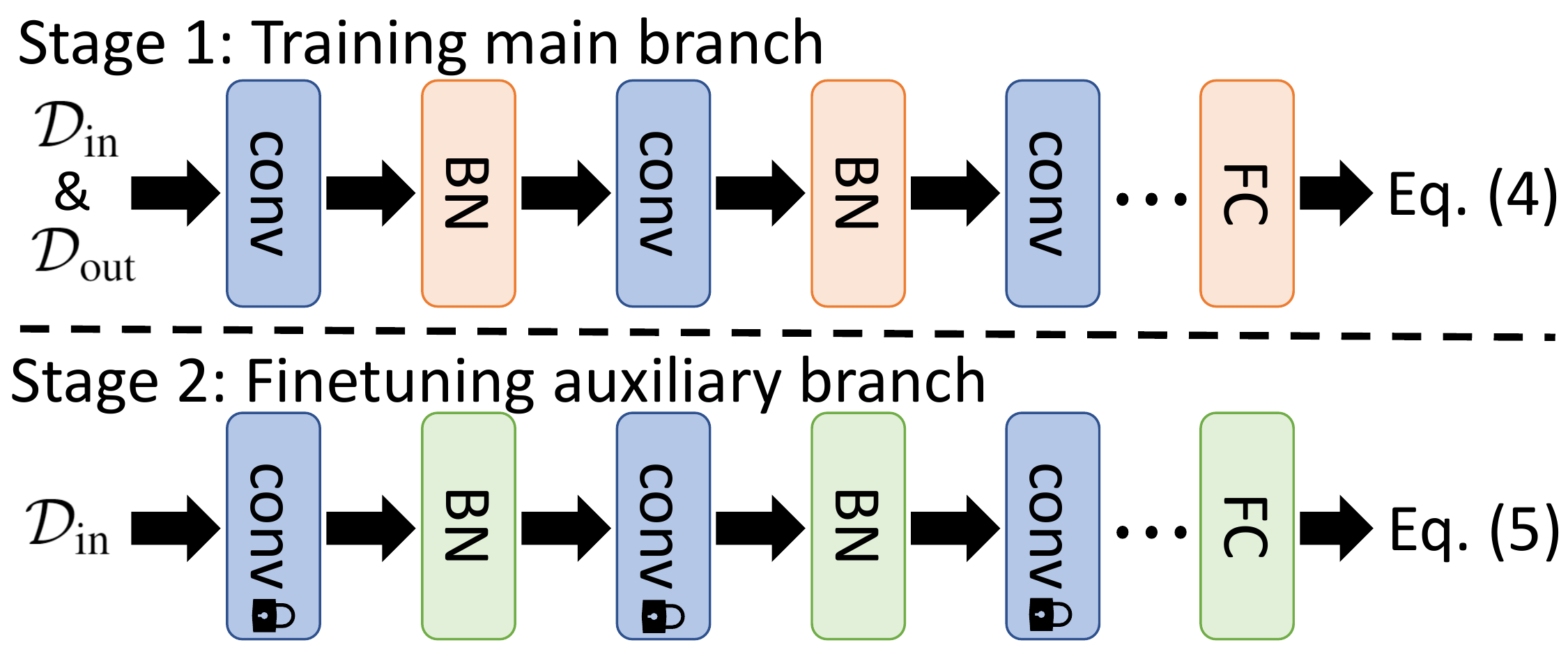} \vspace{-0.5em}\\
     \small{(a) Training} \\
     \includegraphics[width=\columnwidth]{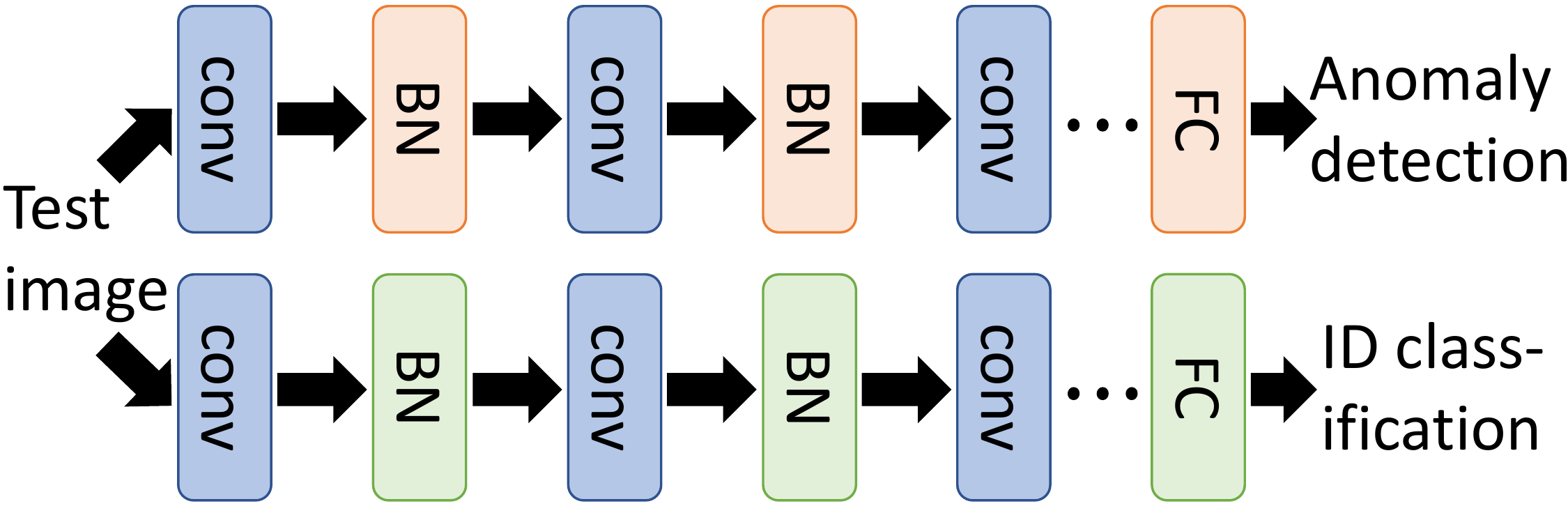} \vspace{-0.5em}\\
     \small{(b) Testing} \\
\end{tabular}
}
\vspace{-1.5em}
\caption{The framework of auxiliary branch finetuning (ABF) at (a) training and (b) test time.
Orange layers are trained in stage one. Green layers are finetuned in stage two, using the weights learned in stage one as initialization. Blue layers are trained in stage one and kept fixed (with the lock icon) in stage two.
}
\label{fig:abf}
\end{figure}

The overall framework of auxiliary branch finetuning is shown in Figure \ref{fig:abf}.
We add an auxiliary counterpart for each BN layer and the classification layer in the model.
The training process consists of two stages. 
In the first stage, we train the main branch using both in-distribution and OOD data, by minimizing Eq.~\eqref{eq:final}. 
In the second stage, we finetune the auxiliary BN and classification layers, while fixing all the other layers, by minimizing the logits-adjustment (LA) cross-entropy loss \cite{menon2020long} only on in-distribution data\footnote{
Naively combining LA with OE has been shown ineffective in Table \ref{tab:benchmark-ood-lt}. 
} for a few iterations:
\begin{equation}
\mathbb{E}_{x\sim\din}[\Lin^{LA}(x)].
\label{eq:abf}
\end{equation}
During test time, we forward each test image twice, using the main and auxiliary branches respectively. 
The outputs from the main branch are used for anomaly detection, while those from the second branch are used for in-distribution classification.

The final PASCL framework is summarized in Algorithm \ref{alg:algo}.

\vspace{-0.5em}
\begin{algorithm}[H]
   \caption{Partial and Asymmetric Supervised Contrastive Learning (PASCL)}
   \label{alg:algo}
\begin{algorithmic}
   \STATE {\bfseries Input:} in-distribution training set  $\din$, OOD training set  $\dout$, main branch training iteration $n_1$, auxiliary branch finetuning iteration $n_2$.
   \STATE \#Stage 1: Train main branch.
   \FOR{$i=1$ {\bfseries to} $n_1$}
   \STATE Sample a batch of in-distribution and OOD training samples.
   \STATE Update the main branch model by minimizing Eq. \eqref{eq:final}
   \ENDFOR
   \STATE \#Stage 2: Finetune auxiliary branch.
   \STATE Fix all layers except the auxiliary BN and classification layers in the model. 
   \FOR{$i=1$ {\bfseries to} $n_2$}
   \STATE Sample a batch of in-distribution training samples.
   \STATE Update the auxiliary BN and classification layers by minimizing Eq. \eqref{eq:abf}. 
   \ENDFOR
\end{algorithmic}
\end{algorithm}
\vspace{-1em}

\section{Experiments}

\subsection{Experiment Settings}
\label{sec:settings}

\begin{table}[t]
\centering
\vspace{-0.5em}
\caption{Results on CIFAR10-LT using ResNet18. The best results are shown in bold. Mean and standard deviation over six random runs are reported for OE and our method.
``Average'' means the results averaged across six different $\douttest$ sets.
}
\resizebox{1.05\linewidth}{!}{
\begin{tabular}{c}
\small{\makecell{(a) OOD detection results and \\in-distribution classification results in terms of ACC95.}} \\
\resizebox{1\linewidth}{!}{
\begin{tabular}{ c|c|ccc|c }
\toprule
$\douttest$ & \textbf{Method} & \textbf{AUROC ($\uparrow$)} & \textbf{AUPR ($\uparrow$)} & \textbf{FPR95 ($\downarrow$)} & \textbf{ACC95 ($\uparrow$)} \\
\midrule
\midrule
\multirow{2}{*}{Texture} 
& OE & 92.59 $\pm$	0.42 & 83.32 $\pm$ 1.67 &	25.10 $\pm$ 1.08 & 84.52 $\pm$ 0.76 \\
& \ours & \textbf{93.16} $\pm$ 0.37 & \textbf{84.80} $\pm$ 1.50 & \textbf{23.26} $\pm$ 0.91 & \textbf{85.86} $\pm$ 0.72\\
\midrule
\midrule
\multirow{2}{*}{SVHN} 
& OE & 95.10 $\pm$ 1.01 & 97.14 $\pm$ 0.81 & 16.15 $\pm$ 1.52 & 81.33 $\pm$ 0.81 \\
& \ours & \textbf{96.63} $\pm$ 0.90 & \textbf{98.06} $\pm$ 0.56 & \textbf{12.18} $\pm$ 3.33 & \textbf{82.72} $\pm$ 1.51\\
\midrule
\midrule
\multirow{2}{*}{CIFAR100} 
& OE & 83.40 $\pm$	0.30 & 80.93 $\pm$ 0.57 & \textbf{56.96}	$\pm$ 0.91 & \textbf{94.56} $\pm$ 0.57 \\
& \ours & \textbf{84.43} $\pm$ 0.23 & \textbf{82.99} $\pm$ 0.48 & 57.27 $\pm$ 0.88 & 94.48 $\pm$ 0.31 \\
\midrule
\midrule
\multirow{2}{*}{\makecell{Tiny\\ImageNet}} 
& OE & 86.14 $\pm$ 0.29 & 79.33 $\pm$ 0.65 & 47.78 $\pm$ 0.72 & 91.19 $\pm$ 0.33 \\
& \ours & \textbf{87.14} $\pm$ 0.18 & \textbf{81.54} $\pm$0.38 & \textbf{47.69} $\pm$ 0.59 & \textbf{91.20} $\pm$ 0.35 \\
\midrule
\midrule
\multirow{2}{*}{LSUN} 
& OE & 91.35 $\pm$	0.23 & 87.62 $\pm$ 0.82 & 27.86	$\pm$ 0.68 & 85.49 $\pm$ 0.69 \\
& \ours & \textbf{93.17} $\pm$ 0.15 & \textbf{91.76} $\pm$ 0.53 & \textbf{26.40} $\pm$ 1.00 & \textbf{86.67} $\pm$ 0.90 \\
\midrule
\midrule
\multirow{2}{*}{Places365} 
& OE & 90.07 $\pm$ 0.26 & 95.15 $\pm$ 0.24 & 34.04 $\pm$ 0.91 & 87.07 $\pm$ 0.53 \\
& \ours & \textbf{91.43} $\pm$ 0.17 & \textbf{96.28} $\pm$ 0.14 & \textbf{33.40} $\pm$ 0.88 & \textbf{87.87} $\pm$ 0.71 \\
\midrule
\midrule
\multirow{2}{*}{Average} 
& OE & 89.77 $\pm$ 0.27 & 87.25 $\pm$ 0.61 & 34.65 $\pm$ 0.46 & 87.36 $\pm$ 0.51 \\
& \ours &  \textbf{90.99} $\pm$  0.19 & \textbf{89.24} $\pm$ 0.34 & \textbf{33.36} $\pm$ 0.79 & \textbf{88.13} $\pm$ 0.56\\
\midrule
\bottomrule
\end{tabular}
} 
\vspace{0.5em}
\\
\small{\makecell{(b) In-distribution classification results in terms of ACC@FPR$n$.}} \\
\resizebox{0.9\linewidth}{!}{
\begin{tabular}{ c|cccc }
\toprule
\multirow{2}{*}{\textbf{Method}}  & \multicolumn{4}{c}{\textbf{ACC@FPRn ($\uparrow$)}}   \\
& 0 & 0.001 & 0.01 & 0.1 \\
\midrule
\midrule
OE & 73.84 $\pm$ 0.77 & 73.90	$\pm$ 0.77 & 74.46 $\pm$ 0.81 & 78.88 $\pm$ 0.66 \\
\ours & \textbf{77.08} $\pm$ 1.01 & \textbf{77.13} $\pm$ 1.02 & \textbf{77.64} $\pm$ 0.99 & \textbf{81.96} $\pm$ 0.85 \\
\bottomrule
\end{tabular}
}
\vspace{0.5em}
\\
\small{(c) Comparison with other methods.} \\
\resizebox{1\linewidth}{!}{
\begin{tabular}{ c|c|ccc|c }
\toprule
$\douttest$ & \textbf{Method} & \textbf{AUROC ($\uparrow$)} & \textbf{AUPR ($\uparrow$)} & \textbf{FPR95 ($\downarrow$)} & \textbf{ACC ($\uparrow$)} \\
\midrule
\midrule
\multirow{6}{*}{Average} 
& ST (MSP) & 72.28 & 70.27 & 66.07 & 72.34\\
& OECC & 87.28 & 86.29 & 45.24 & 60.16 \\
& EnergyOE & 89.31 & \underline{88.92} & 40.88 & \underline{74.68} \\
& OE & \underline{89.77} $\pm$ 0.27 & 87.25 $\pm$ 0.61 & \underline{34.65} $\pm$ 0.46 & 73.84 $\pm$ 0.77 \\
& \ours &  \textbf{90.99} $\pm$  0.19 & \textbf{89.24} $\pm$ 0.34 & \textbf{33.36} $\pm$ 0.79 & \textbf{77.08} $\pm$ 1.01 \\
\midrule
\bottomrule
\end{tabular}
} 
\\
\end{tabular}
}
\vspace{-1em}
\label{tab:cifar10-lt-0.01-ResNet18}
\end{table}

\begin{table}[t]
\centering
\vspace{-0.5em}
\caption{Results on CIFAR100-LT using ResNet18. The best results are shown in bold. Mean and standard deviation over six random runs are reported for OE and our method.
``Average'' means the results averaged across six different $\douttest$ sets.}
\resizebox{1.05\linewidth}{!}{
\begin{tabular}{c}
\small{\makecell{(a) OOD detection results and \\in-distribution classification results in terms of ACC95.}} \\
\resizebox{1\linewidth}{!}{
\begin{tabular}{ c|c|ccc|c }
\toprule
$\douttest$ & \textbf{Method} & \textbf{AUROC ($\uparrow$)} & \textbf{AUPR ($\uparrow$)} & \textbf{FPR95 ($\downarrow$)} & \textbf{ACC95 ($\uparrow$)} \\
\midrule
\midrule
\multirow{2}{*}{Texture} 
& OE & \textbf{76.71} $\pm$ 1.20 &	\textbf{58.79} $\pm$ 1.39 &	68.28 $\pm$ 1.53 & 	71.43 $\pm$ 1.58 \\
& \ours & {76.01} $\pm$ 0.66 & {58.12} $\pm$ 1.06 & \textbf{67.43} $\pm$ 1.93 & \textbf{73.11} $\pm$ 1.55 \\
\midrule
\midrule
\multirow{2}{*}{SVHN} 
& OE & 77.61 $\pm$ 3.26 & 86.82 $\pm$ 2.50 & 58.04 $\pm$ 4.82 & 64.27 $\pm$ 3.26 \\
& \ours & \textbf{80.19} $\pm$ 2.19 & \textbf{88.49} $\pm$ 1.59 & \textbf{53.45} $\pm$ 3.60 & \textbf{64.50} $\pm$ 1.87\\
\midrule
\midrule
\multirow{2}{*}{CIFAR10} 
& OE & 62.23 $\pm$ 0.30 & \textbf{57.57} $\pm$ 0.34 & 80.64 $\pm$ 0.98 & \textbf{82.67} $\pm$ 0.99 \\
& \ours & \textbf{62.33} $\pm$ 0.38 & 57.14 $\pm$ 0.20 & \textbf{79.55} $\pm$ 0.84 & {82.30} $\pm$ 1.07\\
\midrule
\midrule
\multirow{2}{*}{\makecell{Tiny\\ImageNet}} 
& OE & 68.04 $\pm$ 0.37 & \textbf{51.66} $\pm$ 0.51 & 76.66 $\pm$ 0.47 & 76.22 $\pm$ 0.61 \\
& \ours & \textbf{68.20} $\pm$ 0.37 & 51.53 $\pm$ 0.42 & \textbf{76.11} $\pm$ 0.80 & \textbf{77.56} $\pm$ 1.15\\
\midrule
\midrule
\multirow{2}{*}{LSUN} 
& OE & 77.10 $\pm$ 0.64 & \textbf{61.42} $\pm$ 0.99 & 63.98 $\pm$ 1.38 &	65.64 $\pm$ 1.03 \\
& \ours & \textbf{77.19} $\pm$ 0.44 & {61.27} $\pm$ 0.72 & \textbf{63.31} $\pm$ 0.87 & \textbf{68.05} $\pm$ 1.24 \\
\midrule
\midrule
\multirow{2}{*}{Places365} 
& OE & 75.80 $\pm$ 0.45 &	\textbf{86.68} $\pm$ 0.38 &	65.72 $\pm$ 0.92 &	67.04 $\pm$ 0.49 \\
& \ours &  \textbf{76.02} $\pm$ 0.21 & 86.52 $\pm$ 0.29 & \textbf{64.81} $\pm$ 0.27 & \textbf{69.04} $\pm$ 0.90 \\
\midrule
\midrule
\multirow{2}{*}{Average} 
& OE & 72.91 $\pm$ 0.68 & 67.16 $\pm$ 0.57 & 68.89 $\pm$ 1.07 & 71.21 $\pm$ 0.84 \\
& \ours & \textbf{73.32} $\pm$ 0.32 & \textbf{67.18} $\pm$ 0.10 & \textbf{67.44} $\pm$ 0.58 & \textbf{72.43} $\pm$ 0.66 \\
\midrule
\bottomrule
\end{tabular}
} 
\vspace{0.5em}
\\
\small{\makecell{(b) in-distribution classification results in terms of ACC@FPR$n$.}} \\
\resizebox{0.9\linewidth}{!}{
\begin{tabular}{ c|cccc }
\toprule
\multirow{2}{*}{\textbf{Method}}  & \multicolumn{4}{c}{\textbf{ACC@FPRn ($\uparrow$)}}   \\
& 0 & 0.001 & 0.01 & 0.1 \\
\midrule
\midrule
OE & 39.04 $\pm$ 0.37  & 39.07 $\pm$ 0.38 & 39.38 $\pm$ 0.38 & 42.40 $\pm$ 0.44 \\
\ours & \textbf{43.10} $\pm$ 0.47 & \textbf{43.12} $\pm$ 0.47 & \textbf{43.39} $\pm$ 0.48 & \textbf{46.14} $\pm$ 0.38 \\
\bottomrule
\end{tabular}
}
\vspace{0.5em}
\\
\small{(c) Comparison with other methods.} \\
\resizebox{1\linewidth}{!}{
\begin{tabular}{ c|c|ccc|c }
\toprule
$\douttest$ & \textbf{Method} & \textbf{AUROC ($\uparrow$)} & \textbf{AUPR ($\uparrow$)} & \textbf{FPR95 ($\downarrow$)} & \textbf{ACC ($\uparrow$)} \\
\midrule
\midrule
\multirow{6}{*}{Average} 
& ST (MSP) & 61.00 & 57.54 & 82.01 & \underline{40.97}  \\
& OECC &  70.38 & 66.87 & 73.15 & 32.93 \\
& EnergyOE & 71.10 & \underline{67.23} & 71.78 & 39.05  \\
& OE & \underline{72.91} $\pm$ 0.68 & 67.16 $\pm$ 0.57 & \underline{68.89} $\pm$ 1.07 & 39.04 $\pm$ 0.37  \\
& \ours & \textbf{73.32} $\pm$ 0.32 & \textbf{67.18} $\pm$ 0.10 & \textbf{67.44} $\pm$ 0.58 & \textbf{43.10} $\pm$ 0.47 \\
\midrule
\bottomrule
\end{tabular}
} 
\end{tabular}
}
\vspace{-1em}
\label{tab:cifar100-lt-0.01-ResNet18}
\end{table}

\begin{table*}[ht]
\vspace{-0.5em}
\centering
\caption{Results on ImageNet-LT using ResNet50. The best and second-best results are bolded and underlined, respectively. Improvements of our method over OE are shown in parentheses.}
\resizebox{\linewidth}{!}{
\begin{tabular}{ c|c|cc|cccc|cccc|cccc }
\toprule
\multirow{2}{*}{$\douttest$} & \multirow{2}{*}{\textbf{Method}} & \multirow{2}{*}{\textbf{AUROC ($\uparrow$)}} & \multirow{2}{*}{\textbf{AUPR ($\uparrow$)}} & \multicolumn{4}{c|}{\textbf{FPR@TPR$n$ ($\downarrow$)}} & \multicolumn{4}{c|}{\textbf{ACC@TPR$n$ ($\uparrow$)}} & \multicolumn{4}{c}{\textbf{ACC@FPR$n$ ($\uparrow$)}}  \\
& & & & 0.98 & 0.95 & 0.90 & 0.80 & 0.98 & 0.95 & 0.90 & 0.80 & 0 & 0.001 & 0.01 & 0.1 \\
\midrule
\midrule
\multirow{6}{*}{\makecell{ImageNet\\-1k-OOD}} 
& ST (MSP) & 53.81	& 51.63	&95.38&	90.15&	83.52&	72.97	&\textbf{96.67}&	\textbf{92.61}&	\textbf{87.43}&	\textbf{77.52}&	\underline{39.65} &	\underline{39.68} &	\underline{40.00} &	\underline{43.18} \\
\cline{2-16}
& OECC & 63.07 & 63.05 & \textbf{93.15}  & \textbf{86.90} & 78.79 & 65.23 & 94.25 & 88.23 & 80.12 & 68.36 & 38.25 & 38.28 & 38.56 & 41.47 \\
\cline{2-16}
& EnergyOE & 64.76&	64.77&	\underline{94.15}&	87.72 &	\underline{78.36}&	\underline{63.71} &	80.18&	74.38&	67.65&	59.68&	38.50&	38.52&	38.72&	40.99 \\
\cline{2-16}
& OE & \underline{66.33}&	\underline{68.29}&	95.11&	88.22&	78.68&	65.28&	95.46&	88.22&	78.68&	65.28&	37.60&	37.62&	37.79&	40.00 \\
\cline{2-16}
& \multirow{2}{*}{\ours} & \textbf{68.00}&	\textbf{70.15}&	94.38 &	\underline{87.53}&	\textbf{78.12}&	\textbf{62.48} & \underline{95.69} & \underline{89.55} &	\underline{80.88} &	\underline{69.60} &	\textbf{45.49}&	\textbf{45.51}&	\textbf{45.62}&	\textbf{47.49} \\
& & (+1.67) & (+1.86) &	(-0.73) & (-0.69) &	(-0.56) & (-2.80) & (+0.23) & (+1.33) & (+2.20) & (+4.32) & (+7.89) & (+7.89) & (+7.83) & (+7.49) \\
\midrule
\bottomrule
\end{tabular}
}
\vspace{-1em}
\label{tab:imagenet-lt-ResNet50}
\end{table*}

\begin{table*}[ht]
\centering
\vspace{-0.5em}
\caption{The importance of each component (i.e., partiality, asymmetry, and the auxiliary branch) in PASCL. Experiments are conducted on CIFAR10-LT and CIFAR100-LT (both with $\rho=100$) with ResNet18. The first row in each block, where no supervised contrastive learning is used, is the OE baseline. 
SVHN is used as $\douttest$. Mean and standard deviation over six random runs are reported.}
\resizebox{1\linewidth}{!}{
\begin{tabular}{c|ccc|ccc|c|cccc }
\toprule
\multirow{2}{*}{$\din$} & \multirow{2}{*}{Asymmetry} & \multirow{2}{*}{Partiality} & \multirow{2}{*}{ABF} & \multirow{2}{*}{\textbf{AUROC ($\uparrow$)}} & \multirow{2}{*}{\textbf{AUPR ($\uparrow$)}} & \multirow{2}{*}{\textbf{FPR95 ($\downarrow$)}} & \multirow{2}{*}{\textbf{ACC95 ($\uparrow$)}} & \multicolumn{4}{c}{\textbf{ACC@FPR$n$ ($\uparrow$)}}  \\
& & & & & & & & 0 & 0.001 & 0.01 & 0.1\\
\midrule
\midrule
\multirow{6}{*}{CIFAR10-LT} & \multicolumn{3}{c|}{No contrastive loss (OE)} & 95.10 $\pm$ 1.01  & 97.14 $\pm$ 0.81 & 16.15 $\pm$ 1.52 & 81.33 $\pm$ 0.81 & 73.84 $\pm$ 0.77 & 73.90 $\pm$ 0.77 & 74.46 $\pm$ 0.81 & 78.88 $\pm$ 0.66\\
& \xmark & \xmark & \xmark & \underline{95.34} $\pm$ 1.58 & \underline{97.30} $\pm$ 1.20 & \underline{15.12} $\pm$ 3.07 & 81.94 $\pm$ 1.28 & 75.03 $\pm$ 1.46 & 75.09 $\pm$ 1.45 & 75.60 $\pm$ 1.44 & 80.02 $\pm$ 1.10 \\
& \xmark & \cmark & \xmark &  95.01 $\pm$ 1.25 & 96.74 $\pm$ 0.78 & 15.31 $\pm$ 4.35 & \underline{82.34} $\pm$ 1.56 & 74.46 $\pm$ 1.80 & 74.52 $\pm$ 1.80 & 75.04 $\pm$ 1.76 & 80.21 $\pm$ 0.99\\
& \cmark & \xmark & \xmark & 94.91 $\pm$ 1.43 & 96.86 $\pm$ 1.47 & 15.57 $\pm$ 1.19 & 82.08 $\pm$ 0.47 & 75.24 $\pm$ 0.99 & 75.29 $\pm$ 0.98 & 75.77 $\pm$ 0.98 & 79.85 $\pm$ 0.77 \\
& \cmark & \cmark & \xmark & \textbf{96.63} $\pm$ 0.90 & \textbf{98.06} $\pm$ 0.56 & \textbf{12.18} $\pm$ 3.33 & 81.70 $\pm$ 1.21 & \underline{76.20} $\pm$ 0.79 & \underline{76.26} $\pm$ 0.79 & \underline{76.85} $\pm$ 0.81 & \underline{81.07} $\pm$ 0.58  \\
& \cmark & \cmark & \cmark & \textbf{96.63} $\pm$ 0.90 & \textbf{98.06} $\pm$ 0.56 & \textbf{12.18} $\pm$ 3.33 & \textbf{82.72} $\pm$ 1.51 & \textbf{77.08} $\pm$ 1.01 & \textbf{77.13} $\pm$ 1.02 & \textbf{77.64} $\pm$ 0.99 & \textbf{81.96} $\pm$ 0.85 \\
\midrule
\midrule
\multirow{6}{*}{CIFAR100-LT} & \multicolumn{3}{c|}{No contrastive loss  (OE)} & 77.61 $\pm$ 3.26 & 86.82 $\pm$ 2.50 & 58.04 $\pm$ 4.82 & 64.27 $\pm$ 3.26 & 39.04 $\pm$ 0.37 & 39.07 $\pm$ 0.38 & 39.38 $\pm$ 0.38 & 42.40 $\pm$ 0.44 \\
& \xmark & \xmark & \xmark & 78.05 $\pm$ 2.12 & 87.18 $\pm$ 0.87 & 59.10 $\pm$ 5.03 & \textbf{66.44} $\pm$ 3.90 & 40.21 $\pm$ 0.43 & 40.25 $\pm$ 0.43 & 40.56 $\pm$ 0.45 & 43.71 $\pm$ 0.42 \\
& \xmark & \cmark & \xmark & 79.46 $\pm$ 1.83 & \underline{88.01} $\pm$ 1.90 & 54.59 $\pm$ 3.34 & 63.86 $\pm$ 2.52 & 40.24 $\pm$ 0.53 & 40.28 $\pm$ 0.53 & 40.60 $\pm$ 0.55 & \underline{43.93} $\pm$ 0.57\\
& \cmark & \xmark & \xmark & \underline{79.54} $\pm$ 2.38 & 87.68 $\pm$ 1.51 & \underline{54.27} $\pm$ 3.69 & 63.33 $\pm$ 2.87 & 40.00 $\pm$ 0.42 & 40.04 $\pm$ 0.41 & 40.36 $\pm$ 0.42 & 43.60 $\pm$ 0.42 \\
& \cmark & \cmark & \xmark & \textbf{80.19} $\pm$ 2.19 & \textbf{88.49} $\pm$ 1.59 & \textbf{53.45} $\pm$ 3.60 & 63.10 $\pm$ 1.87 & \underline{40.33} $\pm$ 0.20 & \underline{40.36} $\pm$ 0.20 & \underline{40.66} $\pm$ 0.18 & 43.79 $\pm$ 0.22 \\
& \cmark & \cmark & \cmark & \textbf{80.19} $\pm$ 2.19 & \textbf{88.49} $\pm$ 1.59 & \textbf{53.45} $\pm$ 3.60 & \underline{64.50} $\pm$ 1.87 & \textbf{43.10} $\pm$ 0.47 & \textbf{43.12} $\pm$ 0.47 & \textbf{43.39} $\pm$ 0.48 & \textbf{46.14} $\pm$ 0.38 \\
\midrule
\bottomrule
\end{tabular}
}
\label{tab:abla-asymmetric}
\end{table*}

\paragraph{In-distribution training and test sets ($\din$, $\dintest$)}
We use three popular long-tailed image classification datasets, CIFAR10-LT, CIFAR100-LT \cite{cao2019learning}, and ImageNet-LT \cite{liu2019large}, as the in-distribution training data (i.e., $\din$).
Following \citet{menon2020long}, we use the default imbalance ratio $\rho=100$ on CIFAR10-LT and CIFAR100-LT, and we conduct
ablation studies on different $\rho$ in Section \ref{sec:ablation}.
We use the original CIFAR10 and CIFAR100 test sets and the ImageNet validation set as the in-distribution test sets (i.e., $\dintest$).

\vspace{-1em}
\paragraph{OOD training set $\dout$}
For experiments on CIFAT10-LT and CIFAR100-LT, we use TinyImages80M \cite{torralba200880} as the unlabeled OOD training images (i.e., $\dout$) following \cite{hendrycks2018deep,liu2020energy}.
For experiments on ImageNet-LT, we construct our own $\dout$ named ImageNet-Extra. 
Specifically, ImageNet-Extra contains $517,711$ images belonging to $500$ classes randomly sampled from ImageNet-22k \cite{deng2009imagenet} but not overlapping with the $1,000$ in-distribution classes in ImageNet-LT.

\vspace{-1em}
\paragraph{OOD test set $\douttest$}
For experiments on CIFAR10-LT and CIFAR100-LT, we use the recently published semantically coherent out-of-distribution detection (SC-OOD) benchmark datasets \cite{yang2021semantically} as OOD test sets $\douttest$.
Specifically, for experiments on CIFAR10-LT, we use Textures \cite{cimpoi2014describing}, SVHN \cite{netzer2011reading}, CIFAR100, Places365 \cite{zhou2017places}, LSUN \cite{yu2015lsun}, and Tiny ImageNet \cite{letiny} as OOD evaluation datasets $\douttest$; for experiments on CIFAR100-LT, we use Textures, SVHN, CIFAR10, Places365, LSUN, and Tiny ImageNet as $\douttest$.\footnote{According to \citet{yang2021semantically}, not all images in those datasets are OOD images with respect to CIFAR10 and CIFAR100. We use the benchmarks released by \citet{yang2021semantically} where each test image is categorized into either $\dintest$ or $\douttest$ according to their ground-truth semantic meaning.}
For experiments on ImageNet-LT, we construct ImageNet-1k-OOD as $\douttest$. Specifically, ImageNet-1k-OOD contains $50,000$ OOD test images from $1,000$ classes randomly selected from ImageNet-22k (with $50$ images in each class), which is of the same size as the in-distribution test set. 
The $1,000$ classes in ImageNet-1k-OOD are not overlapped with either the $1,000$ in-distribution classes in ImageNet-LT or the $500$ OOD training classes in ImageNet-Extra.

\vspace{-1em}
\paragraph{Evaluation measures}
Following \citet{hendrycks2018deep,mohseni2020self,yang2021semantically}, we use the below evaluation measures:
\vspace{-1em}
\begin{itemize}[leftmargin=*]
\item \textbf{AUROC}: The area under the receiver operating
characteristic curve. AUROC is equivalent to the probability that a positive example has a larger detector score than a negative example.
A perfect detector has $100\%$ AUROC, while a random guess leads to $50\%$ AUROC.
\vspace{-0.5em}
\item \textbf{AUPR}: The area under precision-recall curve. This is also known as the average precision over all recall values.
\vspace{-0.5em}
\item \textbf{FPR@TPR$n$}: The false positive rate (FPR) when $n$ (in percentage) OOD samples have been successfully detected (i.e., when the true positive rate (TPR) is $n$).
Previous papers \cite{hendrycks2016baseline} mainly use the measure \textbf{FPR95}, which is the abbreviation for FPR@TPR$95\%$. They set $n$ to large values such as $95\%$ since they care about the FPR when most OOD samples are successfully detected.
\vspace{-0.5em}
\item \textbf{ACC@TPR$n$}: The classification accuracy on the remaining in-distribution data when $n$ (in percentage) OOD samples have been successfully detected.
\textbf{ACC95} is the abbreviation for ACC@TPR$95\%$.
\vspace{-0.5em}
\item \textbf{ACC@FPR$n$}: The classification accuracy on the remaining in-distribution date when $n$ (in percentage) in-distribution samples are mistakenly detected as OOD samples (i.e., when FPR is $n$). 
\textbf{ACC@FPR$0$}, also abbreviated as \textbf{ACC}, is the accuracy on the entire in-distribution test set.
Previous papers \cite{yang2021semantically} set $n$ to small values such as $1\%$ since they care about the accuracy when few in-distribution samples are falsely detected as OOD.
By definition, the value of ACC@FPR$n$ is irrelevant to the choice of $\douttest$.
\end{itemize}
\vspace{-0.5em}

\vspace{-1em}
\paragraph{Methods in comparison}
As shown in \autoref{tab:benchmark-ood}, among all those OOD detection methods we evaluated, OE has arguably the best OOD detection performance on long-tailed datasets.
Thus we use OE as the main baseline method.
We also compare with two more recent methods:  EnergyOE \cite{liu2020energy} and OECC \cite{PAPADOPOULOS2021138}.
We also show results of MSP on standard training (ST) models.

\vspace{-1em}
\paragraph{Models}
For experiments on CIFAR10 and CIFAR100, we use the standard ResNet18 \cite{he2016deep} following \citet{yang2021semantically}. 
For experiments on ImageNet, we use ResNet50 \cite{he2016deep}.
More details on experimental settings (e.g., learning rate, etc.) are in Appendix \ref{sec:appx-settings}.

\vspace{-0.5em}
\subsection{Main Results}
\label{sec:results}

Results on CIFAR10-LT, CIFAR100-LT, and ImageNet-LT are shown in Table \ref{tab:cifar10-lt-0.01-ResNet18}, \ref{tab:cifar100-lt-0.01-ResNet18}, and \ref{tab:imagenet-lt-ResNet50}, respectively.
Table \ref{tab:cifar10-lt-0.01-ResNet18} and \ref{tab:cifar100-lt-0.01-ResNet18} both contain three sub-tables: In sub-table (a), we show AUROC, AUPR, FPR95 and ACC95 on each $\dout$ and also the average value on all six $\douttest$ datasets, since these four measures may vary among different $\douttest$.
In sub-table (b), we show ACC@FPR$n$ with different $n$ values, which remains the same regardless of the choice of $\douttest$.
The results of other baseline methods (in terms of both anomaly detection and in-distribution classification) are summarized in sub-table (c) due to space limit. 
The results on ImageNet-LT can be reported within one table, since there is only one $\douttest$.

As we can see, our method outperforms OE and also other baseline methods by a considerable margin. 
For example, 
on CIFAR10-LT, our method achieves $1.22\%$ higher average AUROC, $1.99\%$ higher average AUPR, $1.29\%$ lower average FPR95, $0.77\%$ higher average ACC95, and $3.24\%$ higher in-distribution accuracy than OE. 
On CIFAR100-LT, our method achieves $1.45\%$ lower average FPR95, $1.22\%$ higher average ACC95 and $4.06\%$ higher in-distribution accuracy than OE. 
On ImageNet-LT, our method acheives $1.67\%$ higher AUROC, $1.86\%$ higher AUPR, $1.33\%$ higher average ACC95, and $7.89\%$ higher in-distribution accuracy than OE. 

\vspace{-1em}
\paragraph{Simultaneously improving FPR95 and ACC95}
Previous work has shown it extremely challenging to simultaneously improve FPR95 and ACC95 \cite{mohseni2020self}. The reason is simple: When the FPR95 is high, many in-distribution samples are falsely detected as OOD, including those hard or corner-case in-distribution samples. In result, the remaining in-distribution samples are mostly easy to classify, leading to high ACC95.
We highlight that our method simultaneously achieves better FPR95 and ACC95 than OE. This crucial improvement shows our method learns better features for both the anomaly detection and in-distribution classification tasks.

\vspace{-1em}
\paragraph{Improvements on head and tail in-distribution classes}
\begin{wraptable}[8]{r}{0.6\linewidth}
\centering
\vspace{-2em}
\caption{The results on head and tail classes of ImageNet-LT. The improvements of our method over OE are shown in parentheses.}
\resizebox{1\linewidth}{!}{
\begin{tabular}{ c|cc }
\toprule
\multirow{2}{*}{\textbf{Method}} & \multicolumn{2}{c}{\textbf{ACC ($\uparrow$)}} \\
&  Head classes & Tail classes \\
\midrule
\midrule
OE & 54.29 & 20.90 \\
\ours & 54.73 (+0.44) & 36.26 (+15.36) \\
\midrule
\bottomrule
\end{tabular}
} 
\vspace{-1em}
\label{tab:classwise}
\end{wraptable}
In Table \ref{tab:classwise}, we show the improvements of our method over OE on head and tail in-distribution classes separately. 
As we can see, our method mainly benefits the tail classes.

\subsection{Ablation Study}
\label{sec:ablation}

\paragraph{On partiality, asymmetry, and auxiliary branch finetuning}
As introduced in Section \ref{sec:solutions}, there are three building blocks in our PASCL framework: (i) partiality and (ii) asymmetry in supervised contrastive learning and the (iii) auxiliary branch finetuning (ABF).
In \autoref{tab:abla-asymmetric}, we report the results of ablation studies on the three components using both CIFAR10-LT and CIFAR100-LT, to show the importance of each one.
Here we use SVHN as $\douttest$ since our method has relatively large performance gain over OE on SVHN, making it easier to identify where the performance gain come from. 
First, on both datasets, simply applying the original supervised contrastive learning onto the OE framework (i.e., the second rows in each block in Table \ref{tab:abla-asymmetric}) only brings marginal improvements over the OE baseline (i.e., the first rows in each block in Table \ref{tab:abla-asymmetric}), especially in terms of AUROC and AUPR. 
Second, partiality and asymmetry should always be used together for better performance.
For example, on CIFAR10-LT, neither asymmetric nor partiality brings benefit over the original supervised contrastive learning when applied alone, while great improvements are made when they are used together. 
Third, the auxiliary branch finetuning can significantly boot in-distribution classification results (i.e., the last row in each block in Table \ref{tab:abla-asymmetric}).

\vspace{-1em}
\paragraph{On the percentage of tail classes}

As mentioned in Section \ref{sec:solutions}, we apply contrastive loss on tail-class samples to push them away from OOD samples. A key hyper-parameter here is the threshold to define the separation of head and tail classes: the $k$ (in percentage) classes with the least training samples are defined as tail classes.
Ablation study on $k$ is provided in \autoref{tab:abla-tail}.
As we can see, on both CIFAR10-LT and CIFAR100-LT, the best results are achieved around $k=50\%$ (the default value in our experiments).
The results at $k=50\%$ are considerably better than those at $k=100\%$ (without partiality in supervised contrastive learning) and $k=0\%$ (without supervised contrastive learning, i.e., the OE baseline), showing the importance of our partiality design in supervised contrastive learning. 
Also, the performance of our method is stable with respect to $k$ within a large range (e.g, $k \in [40\%,60\%]$).

\begin{table}[ht]
\centering
\vspace{-1em}
\caption{Ablation study on the percentage ($k$) of tail classes. Experiments are conducted on CIFAR10/100-LT (both with $\rho=100$) using ResNet18. The first row ($k=100\%$) means to apply the asymmetric supervised contrastive learning on all in-distribution training samples (without partiality). The last row ($k=0\%$) is the OE baseline where no supervised contrastive learning is used. 
Average results over six $\douttest$ in SC-OOD benchmarks are reported.}
\resizebox{0.9\linewidth}{!}{
\begin{tabular}{c|c|ccc }
\toprule
$\din$ & $k$ & \textbf{AUROC ($\uparrow$)} & \textbf{AUPR ($\uparrow$)} & \textbf{FPR95 ($\downarrow$)}  \\
\midrule
\midrule
\multirow{5}{*}{CIFAR10-LT} & $100\%$ & 89.73 & 87.05 & 33.05 \\
\cmidrule{2-5}
& $60\%$ & 90.44 & 88.51 & 32.90   \\
& $50\%$ & \textbf{91.10} & \textbf{89.01} & \textbf{32.63}   \\
& $40\%$ & 90.66 & 88.90 & 33.30 \\
\cmidrule{2-5}
& \makecell{$0\%$ (OE)} & 89.45	& 86.85 & 35.21  \\
\midrule
\midrule
\multirow{6}{*}{CIFAR100-LT} & $100\%$ & 72.95 & 66.36 & 67.86 \\
\cmidrule{2-5}
& $60\%$ & 73.24 & 67.26 & 67.59 \\
& $50\%$ & \textbf{73.54} & \textbf{67.29} & \textbf{66.92} \\
& $40\%$ & 72.76 & 67.07 & 68.37 \\
\cmidrule{2-5}
& \makecell{$0\%$ (OE)} &  72.62 & 66.73 & 68.69 \\
\midrule
\bottomrule
\end{tabular}
}
\label{tab:abla-tail}
\end{table}

\vspace{-1em}
\paragraph{On auxiliary branch finetuning layers}
As described in Section \ref{sec:solutions}, auxiliary branch finetuning updates all BN layers and the classification (CLF) layer. 
In this section, 
we experiment with the following choices of finetuned layers: none of the layers (denoted as None), all BN layers (denoted as BN), the classification layer (denoted as CLF), all BN layers and the classification layer (denoted as BN \& CLF), and every layer in the model (denoted as All). The results are shown in Table \ref{tab:ablation_abf}.
As we can see, finetuning all BN layers or the classification layer both helps improve accuracy. Finetuning BN and classification layers simultaneously (our default setting) achieves the best accuracy, even outperforming finetuning all layers. 
A likely explanation is that the convolutional layers from stage one have learned features from OOD training data that are transferable to in-distribution data.
Also, auxiliary branch finetuning  (on BN \& CLF layers) adds only a tiny overhead on model size. 

More experimental results (e.g., ablation study on imbalance ratio, model structure, etc.) can be be found in Appendix \ref{sec:appx-results}.

\begin{table}[ht]
\centering
\caption{Ablation study on auxiliary branch finetuning layers using ImageNet-LT and ResNet50. Model size in Million Bytes (MBs).}
\resizebox{0.95\linewidth}{!}{
\begin{tabular}{ c|ccccc }
\toprule
Layers & None & BN & CLF & BN \& CLF & All \\
\midrule
ACC (\%) & 41.32 & 43.42 & 44.57 & \textbf{45.49} & 45.12 \\
Model size (MB) & \textbf{94.11} & 94.30 & 94.20 &  94.38 & 188.23 \\
\bottomrule
\end{tabular}
} 
\label{tab:ablation_abf}
\end{table}

\section{Summary}
To address the new challenging problem of OOD detection in long-tailed recognition, we proposd a novel PASCL framework to explicitly encourage the model to distinguish tail-class in-distribution samples from OOD samples. 
The core idea of PASCL is that head-class in-distribution, tail-class in-distribution, and OOD samples play different roles in OOD detection under long-tailed recognition and thus should be treated differently in contrastive learning.
Experiments on long-tailed image classification datasets
demonstrated the empirical effectiveness of PASCL.

\vspace{-0.5em}
\section*{Acknowledgement}
\vspace{-0.5em}
Z.W. is supported by the U.S. Army Research Laboratory Cooperative Research Agreement W911NF17-2-0196 (IOBT REIGN) and an Amazon Research Award.

\bibliography{reference}
\bibliographystyle{icml2022}

\clearpage
\appendix

\section{More Details on Experimental Settings}
\label{sec:appx-settings}
In this section, we provide more details on experimental settings in addition to those in Section \ref{sec:settings}.
For experiments on CIFAR10-LT and CIFAR100-LT, we train the main branch (i.e., stage 1 in Algorithm \ref{alg:algo}) for $200$ epochs using Adam \cite{kingma2014adam} optimizer with initial learning rate $1 \times 10^{-3}$ and batch size $256$. We decay the learning rate to $0$ using a cosine annealing learning rate scheduler \cite{loshchilov2016sgdr}. 
For auxiliary branch finetuning (i.e., stage 2 in Algorithm \ref{alg:algo}), we finetune the auxiliary branch for $3$ epochs using Adam optimizer with initial learning rate $5\times 10^{-4}$. Other hyper-parameters are the same as in main branch training. 
For experiments on ImageNet-LT, we follow the settings in \cite{wang2020long}. Specifically, we train the main branch for $100$ epochs using SGD optimizer with initial learning rate $0.1$ and batch size $256$. We decay the learning rate by a factor of $10$ at epoch $60$ and $80$. 
For auxiliary branch finetuning , we finetune the auxiliary branch for $3$ epochs using SGD optimizer with initial learning rate $0.01$, which is decayed by a factor of $10$ after each finetune epoch. 
On all datasets, we set $\tau=0.1$ following \cite{khosla2020supervised}, $\lambda_1=0.5$ following \cite{hendrycks2018deep}, and empirically set $\lambda_2=0.1$ for PASCL.
The total training epochs are kept the same across all compared methods. 
For other hyper-parameters in the baseline methods, we use the suggested values in the original papers.

\section{Feature Visualization on CIFAR10 and CIFAR10-LT}
\label{sec:appx-vis}

\begin{figure}[ht]

\centering
\includegraphics[width=0.8\columnwidth]{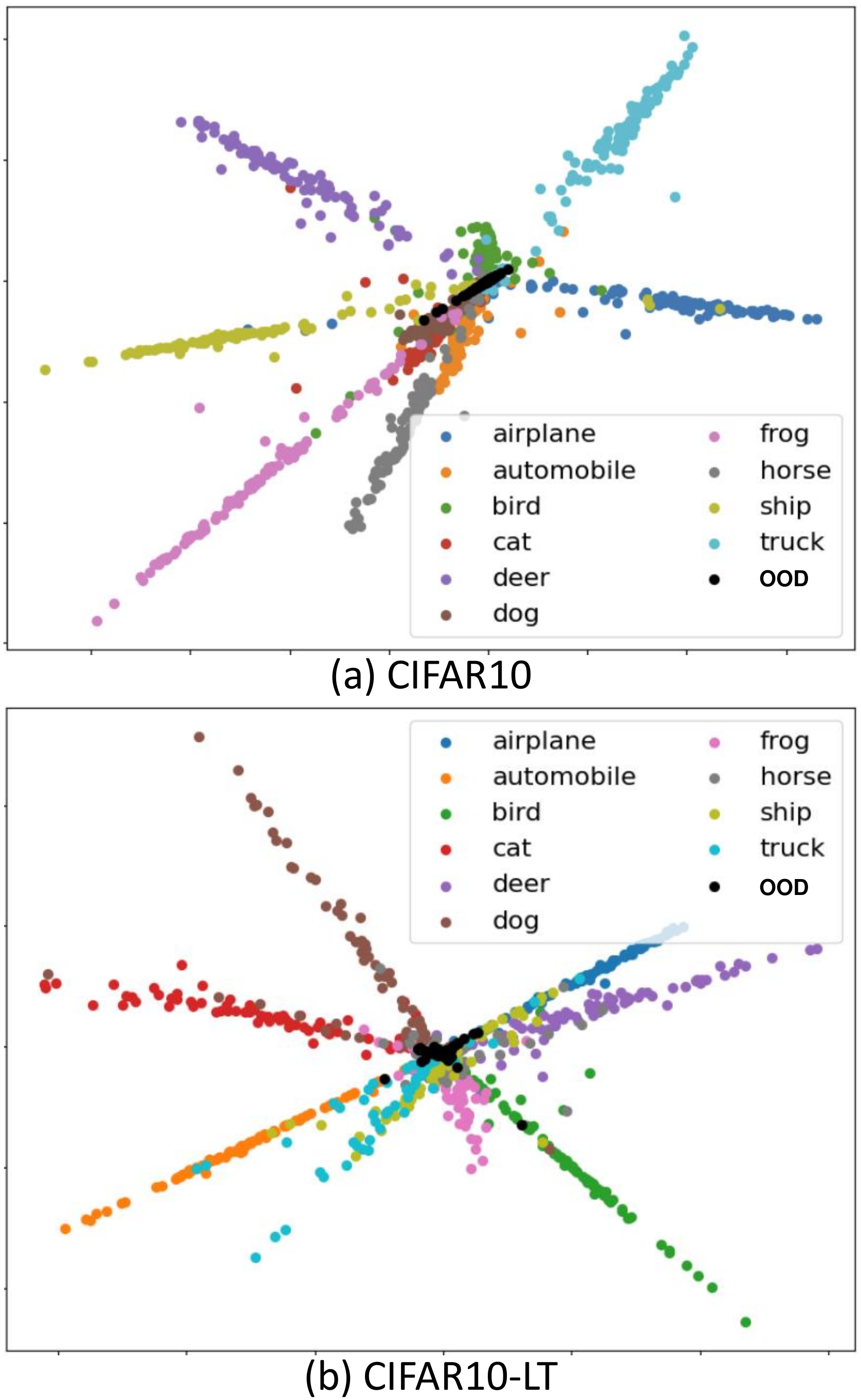}
\vspace{-1em}
\caption{Visualization of in-distribution and OOD samples in the feature space.
We visualize the penultimate layer features of ResNet18 trained by OE on (a) CIFAR10 and (b) CIFAR10-LT, using the same visualization method as in \cite{pang2019rethinking}.
$100$ test samples from each in-distribution class and $100$ OOD samples are visualized.
The model trained on CIFAR10-LT struggles to distinguish tail in-distribution classes like ``horse" (grey) and ``ship" (yellow) from OOD samples (black).
Note that in CIFAR10-LT, the number of training samples in each class follows the alphabetic order: the ``airplane'' class has the most training samples and the ``truck'' class has the least.
}
\vspace{-0.5em}
\label{fig:appx-scatter}
\end{figure}

In Figure \ref{fig:scatter} (in Section \ref{sec:challenge}), we showed the feature distribution of OE trained on CIFAR10-LT. In this section, we compare the feature distribution of OE trained on the balanced CIFAR10 (Figure \ref{fig:appx-scatter} (a)) and CIFAR10-LT (Figure \ref{fig:appx-scatter} (b)).
As we mentioned in Section \ref{sec:challenge}, 
we can see that, on the balanced CIFAR10 dataset, the OOD samples are well separated from all in-distribution classes.
In contrast, when trained on CIFAR10-LT, the features of tail-class in-distribution (e.g., horse, ship) samples and OOD samples heavily overlap, while those of head-class in-distribution (e.g., airplane, automobile, bird) samples are well separated from OOD samples.

\section{More Experimental Results}
\label{sec:appx-results}

\subsection{More Ablation Study}
\paragraph{On imbalance ratio $\rho$}
In Section \ref{sec:results}, we use imbalance ratio $\rho=100$ on both CIFAR10-LT and CIFAR100-LT. In this section, we show that our method can work well under different imbalance ratios. 
Specifically, we conduct experiments on CIFAR10-LT with $\rho=50$. The results are shown in Table \ref{tab:cifar10-lt-0.05-ResNet18}. Our method also outperforms the OE baseline by a considerable margin when $\rho=50$. 

\begin{table}[ht]
\centering
\caption{Results on CIFAR10-LT ($\rho=50$) using ResNet18.}
\resizebox{1\linewidth}{!}{
\begin{tabular}{ c|c|ccc|c }
\toprule
$\douttest$ & \textbf{Method} & \textbf{AUROC ($\uparrow$)} & \textbf{AUPR ($\uparrow$)} & \textbf{FPR95 ($\downarrow$)} & \textbf{ACC ($\uparrow$)} \\
\midrule
\midrule
\multirow{2}{*}{Average} 
& OE & 93.13 & 91.06 & 24.73 & 83.34 \\
& \ours & \textbf{93.94} & \textbf{92.79} & \textbf{22.80} & \textbf{85.44} \\
\midrule
\bottomrule
\end{tabular}
} 
\vspace{-1em}
\label{tab:cifar10-lt-0.05-ResNet18}
\end{table}

\vspace{-0.5em}
\paragraph{On model structures}
In Section \ref{sec:results}, we use the standard ResNet18 as the backbone model. In this section, we show that our method can work well under different model structures, but conducting experiments using the standard ResNet34 \cite{he2016deep}. The results are shown in Table \ref{tab:cifar10-lt-0.01-ResNet34}.  Our method also outperforms the OE baseline by a considerable margin on ResNet34. 

\begin{table}[ht]
\centering
\caption{Results on CIFAR10-LT ($\rho=100$) using ResNet34.}
\resizebox{1\linewidth}{!}{
\begin{tabular}{ c|c|ccc|c }
\toprule
$\douttest$ & \textbf{Method} & \textbf{AUROC ($\uparrow$)} & \textbf{AUPR ($\uparrow$)} & \textbf{FPR95 ($\downarrow$)} & \textbf{ACC ($\uparrow$)} \\
\midrule
\midrule
\multirow{2}{*}{Average} 
& OE & 89.86 & 87.28 & 33.66 & 73.39 \\
& \ours & \textbf{91.11} & \textbf{89.28} & \textbf{33.21} & \textbf{75.34} \\
\midrule
\bottomrule
\end{tabular}
} 
\vspace{-1em}
\label{tab:cifar10-lt-0.01-ResNet34}
\end{table}

\paragraph{Ablation study on $\lambda_2$}
The results are shown in \autoref{tab:abla-lambda2}. The performance of our method is stable with respect to different $\lambda_2$ values.

\begin{table}[t]
\centering
\setlength{\tabcolsep}{3pt}
\caption{Ablation study results on $\lambda_2$. CIFAR100-LT is used as in-distribution training set. SVHN is used as OOD test set.}
\resizebox{0.6\linewidth}{!}{
\begin{tabular}{ c|cccc }
\toprule
\textbf{$\lambda_2$} & 0 & 0.05 & 0.1 & 0.5 \\
\midrule
\textbf{AUPR ($\uparrow$)} & 86.82 & 88.12 & 88.49 & 88.46 \\
\bottomrule
\end{tabular}
}
\label{tab:abla-lambda2}
\end{table}

\subsection{Combining OE with the Original Supervised Contrastive Learning on $\din$ or $\din \cup \dout$}
\label{sec:appx-original-scl}

\begin{table}[ht]
\centering
\caption{Combining OE with the original supervised contrastive learning (SCL) on $\din$ and $\din \cup \dout$. Experiments are conducted on CIFAR10/100-LT (both with $\rho=100$) using ResNet18. SVHN is used as $\douttest$.}
\resizebox{1.0\linewidth}{!}{
\begin{tabular}{c|c|ccc }
\toprule
$\din$ & \textbf{Method} & \textbf{AUROC ($\uparrow$)} & \textbf{AUPR ($\uparrow$)} & \textbf{FPR95 ($\downarrow$)}  \\
\midrule
\midrule
\multirow{2}{*}{CIFAR10-LT} & OE + SCL on $\din$ & 95.30 $\pm$ 1.42 & 97.19 $\pm$ 1.02 &  15.08 $\pm$ 2.54 \\
&  OE+SCL on $\din \cup \dout$ & {95.34} $\pm$ 1.58 & {97.30} $\pm$ 1.20 & {15.12} $\pm$ 3.07 \\
\midrule
\midrule
\multirow{2}{*}{CIFAR100-LT} & OE + SCL on $\din$ & 78.01 $\pm$ 1.95 & 87.16 $\pm$ 0.62 & 59.36 $\pm$ 4.52 \\
& OE+SCL on $\din \cup \dout$ & 78.05 $\pm$ 2.12 & 87.18 $\pm$ 0.87 & 59.10 $\pm$ 5.03 \\
\midrule
\bottomrule
\end{tabular}
}
\vspace{-1em}
\label{tab:appx-original-scl}
\end{table}

In Section \ref{sec:solutions}, we mentioned that when combined with OE, the original supervised contrastive learning achieves slightly better OOD detection performance if applied on $\din \cup \dout$ than on $\din$. 
In this section, we provide experimental results to support this. 
Specifically, we compare the two methods on CIFAR10-LT and CIFAR100-LT, and show the results in Table \ref{tab:appx-original-scl}.
As we can see, these two methods achieve almost identical performance, although those on $\din \cup \dout$ are only slightly better.
For that reason, we use the second method (i.e., supervised contrastive learning on $\din \cup \dout$) as ``the original SCL'' baseline in Table \ref{tab:ablation_abf} (i.e., the second rows in each block in Table \ref{tab:ablation_abf}).

\subsection{OLTR Results}
\label{sec:appx-oltr}

As discussed in Section \ref{sec:related-works}, OLTR \cite{liu2019large} focuses on in-distribution classification in long-tailed recognition instead of OOD detection.
Although it achieves better OOD detection performance than naive baselines such as MSP, it is not comparable with state-of-the-art OOD detection methods such as OE. In contrast, our paper focuses on improve OOD detection performance on long-tailed training sets. 
The results of OLTR and other methods (including our PASCL) on ImageNet-LT are compared in Table \ref{tab:appx-oltr}.

\begin{table}[h!]
\centering
\caption{Comparing OLTR with our PASCL and other methods. Experiments are conducted on ImageNet-LT using ResNet50.}
\resizebox{0.8\linewidth}{!}{
\begin{tabular}{c|ccc }
\toprule
\textbf{Method} & \textbf{AUROC ($\uparrow$)} & \textbf{AUPR ($\uparrow$)} & \textbf{FPR95 ($\downarrow$)}  \\
\midrule
\midrule
ST (MSP) & 53.81 & 51.63 & 90.15 \\
OLTR & 55.68 & 54.02 & 90.02 \\
OE & \underline{66.33}&	\underline{68.29} &	\underline{88.22} \\
Ours & \textbf{68.00} &	\textbf{70.15} & \textbf{87.53} \\
\midrule
\bottomrule
\end{tabular}
}
\vspace{-1em}
\label{tab:appx-oltr}
\end{table}

\section{More Related Works}
\label{sec:appx-related}

In this section, we discuss more related works along with those in Section \ref{sec:related-works}.

\paragraph{OOD detection}
ODIN \cite{liang2018enhancing} improved MSP by adding temperature scaling and adversarial attacks to enlarge the differences in MSP between in-distribution and OOD data.
\citet{hein2019relu} theoretically explained why piecewise linear models tend to give high confident predictions on OOD data, and proposed a new robust learning method to prevent such overconfidence issue. 
\citet{meinke2019towards} proposed Certified Certain Uncertainty (CCU), a robust learning method which achieves provably low confidence predictions on OOD samples. 
Self-Supervised Feature Learning (SOFL) \cite{mohseni2020self} proposed to classify all OOD samples into several abstaining classes, where the pseudo labels for OOD training samples are assigned in a self-supervised manner.  
\citet{tang2021codes} proposed to generate Chamfer OOD samples (i.e., the OOD samples that are close to the in-distribution samples), when there is no OOD  samples directly available for training.
\citet{fort2021exploring} observed that large-scale vision transformers achieves significantly better OOD detection performance than CNNs. They further proposed a simple but surprisingly effective OOD detection method based on multi-model transformers.
\citet{zhou2021step} studied a new problem setting where the number of labeled in-distribution samples are limited. 
\citet{wang2021can} studied the OOD detection problem in multi-label classification. 

\paragraph{Long-tailed recognition}
\citet{cui2019class} proposed a loss reweighing method based on the effective numbers of training samples in each class. 
BBN \cite{zhou2020bbn} used two separate branches which focus on learning universal and tail-class-specific features, respectively. 
\citet{xiang2020learning} split the unbalanced training set into several relatively balanced subsets, and then trained a separate model on each subset, whose knowledge is later distilled into a single model.
RIDE \cite{wang2020long} used mixture of experts and a novel dynamic expert routing module to largely boost long-tailed recognition performance.


\end{document}